\documentclass[journal]{IEEEtran}
\IEEEoverridecommandlockouts

\usepackage{graphicx} 
\usepackage{amsmath,graphics,amssymb,epsfig,subfigure,epstopdf}
\usepackage{amsfonts}
\usepackage{calc}
\usepackage{array}
\usepackage{amsthm}
\usepackage{algorithm}
\usepackage{algorithmicx}
\usepackage{algpseudocode} 
\usepackage{float}
\usepackage{multirow}
\usepackage{midfloat}
\usepackage{nicefrac}
\usepackage{color,soul}
\usepackage[mathscr]{eucal}
\usepackage{booktabs}
\usepackage{textcomp}
\usepackage{cite}
\usepackage{colortbl} 
\usepackage{xcolor} 
\usepackage{makecell} 

\newtheorem{lemma}{Lemma}
\newtheorem{remark}{Remark}

\algnewcommand\algorithmicswitch{\textbf{switch}}
\algnewcommand\algorithmiccase{\textbf{case}}
\algnewcommand\algorithmicassert{\texttt{assert}}
\algnewcommand\Assert[1]{\State \algorithmicassert(#1)}%
\algdef{SE}[SWITCH]{Switch}{EndSwitch}[1]{\algorithmicswitch\ #1\ \algorithmicdo}{\algorithmicend\ \algorithmicswitch}%
\algdef{SE}[CASE]{Case}{EndCase}[1]{\algorithmiccase\ #1}{\algorithmicend\ \algorithmiccase}%
\algtext*{EndSwitch}%
\algtext*{EndCase}%

\def\BibTeX{{\rm B\kern-.05em{\sc i\kern-.025em b}\kern-.08em
T\kern-.1667em\lower.7ex\hbox{E}\kern-.125emX}}

\begin{document}

\title{Cooperative Inference for Real-Time 3D Human Pose Estimation in Multi-Device Edge Networks}

\author{Hyun-Ho Choi,~\IEEEmembership{Senior Member,~IEEE}, Kangsoo Kim,~\IEEEmembership{Member,~IEEE}, Ki-Ho Lee, and Kisong Lee,~\IEEEmembership{Senior Member,~IEEE}%

\thanks{H.-H. Choi is with the School of ICT, Robotics \& Mechanical Engineering, Hankyong National University, Anseong 17579, Republic of Korea (e-mail: hhchoi@hknu.ac.kr).}
\thanks{K. Kim is with the Department of Electrical and Software Engineering, Schulich School of Engineering, University of Calgary, AB, T2N 1N4, Canada (e-mail: kangsoo.kim@ucalgary.ca).}
\thanks{K.-H. Lee is with the School of the Electrical Engineering, Chung-Ang University, Seoul 06974, Republic of Korea (e-mail: kiholee79@snu.ac.kr).}
\thanks{K. Lee is with the Department of Information and Communication Engineering, Dongguk University, Seoul 04620, Republic of Korea (e-mail: kslee851105@gmail.com).}
\thanks{This work has been submitted for possible publication in the IEEE Internet of Things Journal. Copyright may be transferred without notice, after which this version may no longer be accessible.}}

\maketitle

\begin{abstract}
Accurate and real-time three-dimensional (3D) pose estimation is challenging in resource-constrained and dynamic environments owing to its high computational complexity. 
To address this issue, this study proposes a novel cooperative inference method for real-time 3D human pose estimation in mobile edge computing (MEC) networks. 
In the proposed method, multiple end devices equipped with lightweight inference models employ dual confidence thresholds to filter ambiguous images.
Only the filtered images are offloaded to an edge server with a more powerful inference model for re-evaluation, thereby improving the estimation accuracy under computational and communication constraints. 
We numerically analyze the performance of the proposed inference method in terms of the inference accuracy and end-to-end delay and formulate a joint optimization problem to derive the optimal confidence thresholds and transmission time for each device, with the objective of minimizing the mean per-joint position error (MPJPE) while satisfying the required end-to-end delay constraint.
To solve this problem, we demonstrate that minimizing the MPJPE is equivalent to maximizing the sum of the inference accuracies for all devices, decompose the problem into manageable subproblems, and present a low-complexity optimization algorithm to obtain a near-optimal solution.
The experimental results show that a trade-off exists between the MPJPE and end-to-end delay depending on the confidence thresholds.
Furthermore, the results confirm that the proposed cooperative inference method achieves a significant reduction in the MPJPE through the optimal selection of confidence thresholds and transmission times, while consistently satisfying the end-to-end delay requirement in various MEC environments.
\end{abstract}



\section{Introduction}
\label{title:intro}
\IEEEPARstart{T}{hree-dimensional} (3D) human pose estimation is widely utilized in various applications, such as intelligent surveillance, augmented reality, human-computer interaction, and autonomous systems, and requires both high accuracy and timeliness to effectively interpret human motion \cite{wang2021deep}. 
Typically, a 3D pose estimation method involves inferring 2D joint coordinates from images captured by multiple cameras and then reconstructing 3D joint coordinates from the extracted 2D joint coordinates, which results in a high computational complexity.
Moreover, 3D joint coordinates must be continuously updated over time to track human movements, thereby increasing the computational demand for real-time processing.
Employing low-complexity algorithms enables real-time operation but may degrade the estimation accuracy; therefore, 3D pose estimation inherently involves a trade-off between accuracy and latency \cite{zhang2023deep}.
This makes achieving accurate and real-time performance in 3D pose estimation challenging, especially in resource-constrained and dynamic environments.

Practical implementations of 3D human pose estimation can be categorized into three approaches: device-centric, server-centric, and cooperative inference.
In \emph{device-centric inference}, multiple end devices individually perform 2D pose estimation, and their inference results, including 2D joint coordinates, are aggregated at a centralized unit (e.g., a server or head device) for fusion into 3D joint coordinates \cite{choi2021mobilehumanpose}.
By distributing the computationally intensive 2D pose estimation process across multiple end devices, this approach can reduce the inference time and minimize the traffic sent to the centralized unit.
However, the inherent computational limitations of these devices make it difficult to achieve high inference accuracy.

In \emph{server-centric inference}, all end devices transmit the captured 2D images to a powerful server that only performs 2D pose estimation on these images and subsequently computes the 3D joint coordinates \cite{chang2021pose}.
In this approach, the high computational capability of the server ensures a low processing time and high inference accuracy. 
However, the end devices must offload raw image data to the server, which may lead to additional communication delays.
To reduce this communication overhead, the paradigm of mobile edge computing (MEC) has been introduced, wherein the server is placed on the edge nodes within the network (i.e., edge server) to bring it closer to the devices \cite{Mach17}.
Nonetheless, the system still encounters significant transmission delays owing to the large volume of data traffic transmitted by multiple devices and channel uncertainties between the devices and the server.

Device- and server-centric inferences have complementary strengths and weaknesses that have led to the design of a cooperative approach.
\emph{Cooperative inference} combines several inference models with different complexities and performances; for example, by equipping end devices with a lightweight inference model (faster but less accurate) while deploying a powerful inference model (more accurate but slower) on the server \cite{Zhao2020Lightweight}.
This strategy enables the devices to perform an initial simple inference and allows the server to execute a more sophisticated re-inference only when necessary, thereby improving the accuracy compared to device-centric inference, while reducing the latency overhead for server-centric inference.

\subsection{Related Studies}
In the field of computer vision (CV), various 3D human pose estimation techniques have been proposed \cite{wang2021deep}.
Traditionally, two-stage methods, which first estimate 2D key points from multiple cameras and then project them into 3D space, have been extensively studied \cite{qiu2019cross} and further extended to handle multi-person scenarios \cite{bultmann2021real}.
Moreover, single-stage methods that directly estimate 3D coordinates from 2D images have been studied for both single- and multi-person scenarios \cite{mehta2017vnect,mehta2020xnect}.
Recently, video-based methods that utilize temporal dependencies from consecutive video frames have been developed \cite{pavllo20193d}, and single-image-based methods that estimate 3D poses from a single image have been studied \cite{du2024real}.
These studies mainly focused on developing accurate pose estimation algorithms; however, they did not consider the computational and communication constraints in practical deployment environments.

In the field of communication networks, numerous studies have been conducted to efficiently perform cooperative inference in practical network environments \cite{zhou2019edge,Shao20,zhao2019deep, enomoro2021learning,Ahmed2021Hawk,Wang23,Choi24CCNC,Choi24ICOIN}.
Architectures, frameworks, and key technologies have been introduced for collaborative deep neural network (DNN) inference among end devices, edge servers, and cloud centers, considering the trade-off between computational costs and communication overheads \cite{zhou2019edge,Shao20}.
\emph{Cascade inference} was proposed for intelligent edge surveillance, in which an end device deploys a simple neural network (NN) model to reduce computation costs, whereas a cloud server utilizes a sophisticated NN model to improve detection accuracy \cite{zhao2019deep,enomoro2021learning}.
A prototype of cooperative inference was also implemented using a Raspberry Pi device and a web server, wherein the device performed only motion detection on captured images and the server conducted object detection on suspicious images received from the device \cite{Ahmed2021Hawk}.
Moreover, an edge-assisted real-time video analytics framework was designed to coordinate resource-constrained cameras with powerful edge servers, and an adaptive offloading algorithm was proposed by dynamically tuning the confidence threshold to reduce the bandwidth usage while maintaining high accuracy \cite{Wang23}.
A recent study also addressed the problem of optimizing confidence thresholds for task offloading from a device to a server to minimize end-to-end latency while ensuring the required accuracy \cite{Choi24CCNC}, which was further extended to the joint optimization of confidence thresholds and wireless resources to minimize energy-delay costs \cite{Choi24ICOIN}.
These studies extensively explored various frameworks and techniques for cooperative inference in practical network environments; however, they primarily considered a simplified cooperation scenario, focusing on a single device and a server.

Recent research has expanded to incorporate cooperation among multiple end devices and edge servers in MEC networks \cite{shao2022task,Wen23,Mei23Energy,Kim2024Distributed,amanatidis2023cooperative}.
A multi-device cooperative inference framework was introduced for multi-view image classification and object recognition, and the extraction and encoding of local features transmitted from distributed devices to an edge server were optimized to minimize communication overhead and latency \cite{shao2022task}.
A multi-device edge computing system was designed in which multiple devices performed radar sensing to obtain multi-view data and offloaded quantized feature vectors to a centralized server for classification tasks \cite{Wen23}.
Furthermore, a joint optimization problem of task offloading and resource allocation was investigated to minimize energy consumption in multi-device MEC systems under the constraints of heterogeneous delay and resource competition \cite{Mei23Energy}.
Similarly, the joint optimization of task offloading and user association was addressed in MEC networks with multiple devices and servers by considering the communication latency, computation latency, and battery usage \cite{Kim2024Distributed}.
From a different perspective, a novel edge-computing architecture was proposed to distribute cooperative tasks from the edge server to multiple end devices for collaborative object detection in the image sets stored on the server \cite{amanatidis2023cooperative}.
Previous studies have focused on optimizing cooperative inference in multi-device MEC networks, considering several CV applications, such as image classification and object detection; however, their frameworks have not yet been applied to 3D human pose estimation.

\subsection{Motivation and Contributions}
Significant advancements have been made in 3D human pose estimation techniques; however, efforts to apply them in practical environments are scarce \cite{qiu2019cross,bultmann2021real,mehta2017vnect,mehta2020xnect,pavllo20193d,du2024real}.
Various cooperation strategies have been investigated for several applications in multi-device MEC networks; however, no studies have addressed problems specific to 3D human pose estimation \cite{shao2022task,Wen23,Mei23Energy,Kim2024Distributed,amanatidis2023cooperative}.
The overall process of 3D human pose estimation\footnote{In this study, the popular two-stage method for 3D human pose estimation is considered \cite{qiu2019cross,bultmann2021real}.}--capturing images from multiple devices, extracting 2D joint coordinates, and fusing them to compute 3D joint coordinates--aligns well with the MEC network structure.
Specifically, the end devices can perform 2D pose estimation on input images and transmit the resulting 2D joint coordinates to the server, where the 3D joint coordinates are calculated.
Alternatively, considering their limited computational power, end devices can offload some input images to the server, which performs the 2D pose estimation task on their behalf and then computes the 3D joint coordinates.
In this cooperative scenario, determining where to perform inference and how to allocate resources for offloading are essential for optimizing system performance.
Therefore, in this study, we propose a novel cooperative inference method for real-time 3D human pose estimation in multi-device MEC networks and derive an optimal cooperative inference and resource allocation strategy that minimizes the mean per-joint position error (MPJPE) while satisfying an end-to-end delay requirement under given computational and communication constraints.

\begin{figure*}[t]
    \centerline{\includegraphics[width=0.9\linewidth]{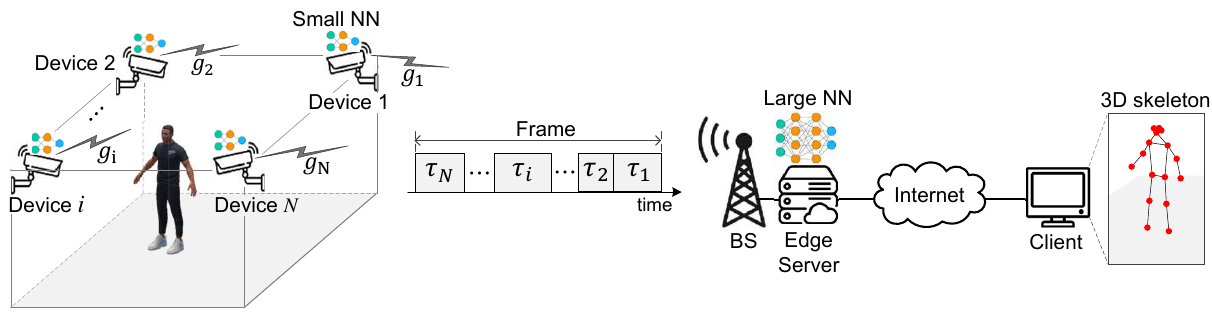}}
    \caption{System model for proposed cooperative inference method.}
    \label{fig1}
\end{figure*}

The detailed contributions of this study are as follows:
\begin{itemize}
    \item A 3D human pose estimation task is applied to an MEC network in which multiple devices and an edge server perform cooperative inference and investigate the joint optimization of cooperative inference and resource allocation to maximize system performance. 
    To the best of our knowledge, this is the first study to address a 3D human pose estimation task in an MEC environment.
    
    \item A novel cooperative inference method is proposed that enables end devices equipped with small NN models to utilize two confidence thresholds to filter ambiguous images with uncertain 2D pose estimation results. 
    Only the filtered images are offloaded to the edge server with a larger NN model for re-inference, thereby reducing the offloading traffic and improving the inference accuracy.
    
    \item The inference accuracy and end-to-end delay performances of the proposed cooperative inference method are numerically analyzed. Moreover, this study proves that minimizing the MPJPE is equivalent to maximizing the sum of the inference accuracies for all devices. 
    Subsequently, a joint optimization problem is solved to determine the optimal confidence thresholds and transmission time for each device, with the aim of maximizing the sum of the inference accuracies while satisfying the required end-to-end delay.
    
    \item Through extensive experiments, this study demonstrates that a trade-off exists between the MPJPE and end-to-end delay depending on the confidence thresholds. 
    Furthermore, the results show that the proposed cooperative inference method, which employs optimal confidence thresholds and transmission times, significantly reduces the MPJPE while satisfying the end-to-end delay requirement. 
\end{itemize}

The remainder of this paper is organized as follows.  
Section \ref{title:prop} describes the system model and the operation of the proposed cooperative inference method in detail.  
Section \ref{title:perf} analyzes the inference accuracy and end-to-end delay performance of the considered methods.
Section \ref{title:opt} formulates a joint optimization problem to minimize the MPJPE while satisfying an end-to-end delay constraint and presents its solution.
Section \ref{title:result} evaluates the proposed method under various scenarios.
Finally, Section \ref{title:con} presents the conclusions of this study.

\section{Proposed Cooperative Inference Method}
\label{title:prop}
Fig. \ref{fig1} shows the system model for the proposed cooperative inference in an MEC network comprising $N$ end devices, a base station (BS), an edge server, and a client.
The end devices (e.g., on-site cameras) are wirelessly connected to the BS, whereas the edge server is directly linked to the BS in close proximity \cite{kekki2018mec}.
The client is connected to the edge server via wired Internet, receives the 3D pose estimation result, and visualizes it as a 3D skeleton for the user.
The end devices and edge server perform cooperative inference for 2D pose estimation using a DNN; then, the server aggregates the 2D pose estimation results and converts them into 3D pose coordinates.
It is assumed that the end devices are equipped with a small NN due to processor and memory limitations, while the edge server is equipped with a large NN without such constraints \cite{zhao2019deep,enomoro2021learning,Ahmed2021Hawk,Wang23}.
The channel gain from device $i$ to the BS is denoted as $g_{i}$. 
Depending on this channel gain and the transmitted data size, the BS determines the transmission time ratio $\tau_{i}$ for each device $i$ within a frame using time-division multiple access.

\begin{figure}[t]
    \centerline{\includegraphics[width=\linewidth]{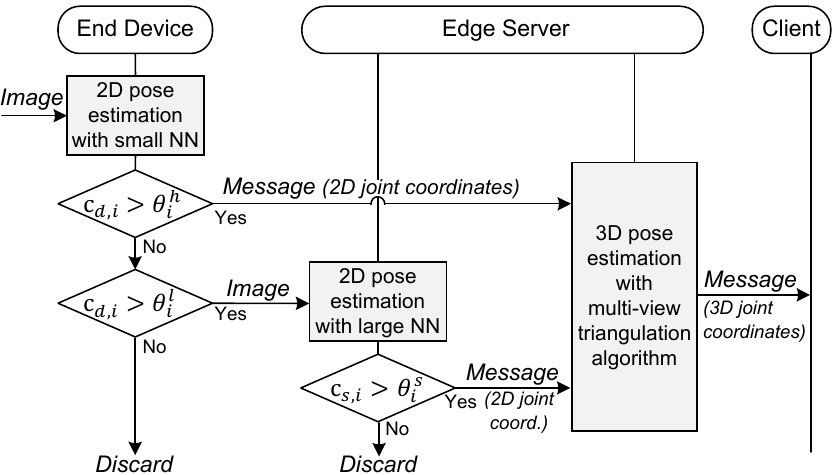}}
    \caption{Operational flow of proposed cooperative inference method.}
    \label{fig2}
\end{figure}

Fig. \ref{fig2} illustrates the operational flow of the proposed cooperative inference method.
Since the end devices use a small NN for 2D pose estimation, their inference results may be relatively less accurate. 
Considering this, the end devices employ two confidence thresholds to filter ambiguous images with uncertain 2D pose estimation results.
To this end, the two confidence thresholds for device $i$ are defined as $\theta_{i}^{l}$ and $\theta_{i}^{h}$, subject to $\theta_{i}^{l} \leq \theta_{i}^{h}$, for $i \!\in\! \mathcal{N} \!=\! \{1,2,\dots,N\}$.
Note that the 2D pose estimation results include the 2D coordinates $(u,v)$ and the corresponding confidence score for each joint of the detected person. 
Thus, we define $c_{d,i}^{j}$ as the confidence score of joint $j$ on device $i$, where $j \!\in\! \mathcal{J} \!=\! \{1,2,\dots,J\}$. 
Without the index $j$, we denote $c_{d,i}$ as the average confidence score of all joints on device $i$.

The end device $i$ performs one of three actions based on the predefined threshold values $\theta_{i}^{l}$ and $\theta_{i}^{h}$, as well as the confidence value $c_{d,i}$ obtained after 2D pose estimation of the input image.
First, if $c_{d,i} \!>\! \theta_{i}^{h}$, device $i$ determines that the 2D pose estimation result is sufficiently confident to be used for 3D pose estimation. 
The device then transmits the resulting message to the edge server containing the 2D joint coordinates $(u,v)$ for each joint of the detected person.
Second, if $\theta_{i}^{l} \!<\! c_{d,i} \!\leq\! \theta_{i}^{h}$, device $i$ judges that the 2D pose estimation result for this image is not sufficiently confident to use. 
Thus, instead of sending the resulting message, the device transmits the original image to the server and requests the server to reprocess the 2D pose estimation using its large NN.
Finally, if $c_{d,i} \!\leq\! \theta_{i}^{l}$, device $i$ determines that the confidence score is too low for this image to be used for 3D pose estimation, concluding that either a non-human object was mistakenly identified as a person or a significant portion of the joints of the person is occluded.
In this case, the device does not send data to the server and the corresponding image is discarded.
That is, if the confidence score is either sufficiently low or high, the device trusts its 2D pose estimation results. 
However, if the confidence score falls within an ambiguous range, the device delegates the 2D pose estimation task to the server. 

When the edge server receives a message from the end devices, it forwards it to the 3D pose estimation block. When it receives an image, it performs 2D pose estimation using its large NN.
Similarly, we denote $c_{s,i}$ as the average confidence score of all joints estimated from the image received by the server from device $i$.
The edge server employs a single confidence threshold $\theta_{i}^{s}$ for each device $i$.
If $c_{s,i} \!>\! \theta_{i}^{s}$, the server determines that its 2D pose estimation result is confident and uses the resulting 2D joint coordinates for 3D pose estimation.
Otherwise (i.e., $c_{s,i} \!\leq\! \theta_{i}^{s}$), the server judges that the 2D estimation result is not useful for 3D pose estimation and discards it.

When the 3D pose estimation block of the edge server receives two or more 2D joint coordinate messages within a given timeframe, it employs a multi-view triangulation algorithm to estimate the 3D joint coordinates.
The triangulation algorithm executes the following calculations \cite{bultmann2021real}. 
For a certain keypoint, its 2D coordinates, estimated from the input image of device $i$, are denoted as $\mathbf{u}_{i} = [u_{i},v_{i}]$, and its estimated 3D coordinates are denoted as $\hat{\mathbf{x}} = [\hat{x},\hat{y},\hat{z}]$.
Moreover, the homogeneous coordinates of $\hat{\mathbf{x}}$ are denoted as $\tilde{\mathbf{x}} = [\hat{x},\hat{y},\hat{z},1]$
The relationship between $\tilde{\mathbf{x}}$ and $\mathbf{u}_{i}$ is then expressed as
\begin{equation}
    \mathbf{A} \tilde{\mathbf{x}} = 0, \label{eq_dlt}
\end{equation}
where the matrix $\mathbf{A}$ is given by
\begin{align}
\mathbf{A} =
\begin{bmatrix}
    u_{1} \mathbf{p}_{1,3} - \mathbf{p}_{1,1} \\
    v_{1} \mathbf{p}_{1,3} - \mathbf{p}_{1,2} \\
    \vdots \\
    u_{N} \mathbf{p}_{N,3} - \mathbf{p}_{N,1} \\
    v_{N} \mathbf{p}_{N,3} - \mathbf{p}_{N,2}
\end{bmatrix}
\in \mathbb{R}^{2N \times 4},
\end{align}
where $\mathbf{p}_{i,k}$ denotes the $k$-th row of the projection matrix $\mathbf{P}_{i}$ of the camera for device $i$, where $\mathbf{P}_i \in \mathbb{R}^{3 \times 4}$.
According to the direct linear transform algorithm \cite{hartley2003multiple}, the triangulation problem \eqref{eq_dlt} is solved by singular value decomposition on $\mathbf{A}$.
The unit singular vector corresponding to the smallest singular value of $\mathbf{A}$ is then taken as the solution for $\tilde{\mathbf{x}}$.
Finally, the 3D coordinates $\hat{\mathbf{x}}$ are obtained as $\hat{\mathbf{x}} = \tilde{\mathbf{x}} / (\tilde{\mathbf{x}})_{4}$ by normalizing $\tilde{\mathbf{x}}$ by its fourth coordinate. 
After performing this calculation for all joint points, the edge server transmits a message containing the estimated 3D joint coordinates to the client.

It is worth noting that the confidence thresholds considered in the proposed cooperative inference method ($\theta_{i}^{l}$, $\theta_{i}^{h}$, $\theta_{i}^{s}$, $\forall i \!\in\! \mathcal{N}$) affect the inference accuracy, offloading decisions, and amount of data transmitted.
Moreover, the transmission time ratio for each device ($\tau_{i}, \forall i \!\in\! \mathcal{N}$) influences the transmission time and the resulting end-to-end delay, as devices share limited radio resources.
Therefore, the optimal selection of $\theta_{i}^{l}$, $\theta_{i}^{h}$, $\theta_{i}^{s}$, and $\tau_{i}$ is critical for improving system performance.

\section{Performance Analysis}
\label{title:perf}
This section presents a numerical analysis of the inference accuracy and end-to-end delay of the proposed cooperative inference method.

\subsection{Analysis of Inference Accuracy}
To analyze the inference accuracy, we separate the average confidence score of device $i$, $c_{d,i}$, into $c_{d,i}^{\mathrm{p}}$ and $c_{d,i}^{\mathrm{n}}$, depending on whether the ground truth of the input image is positive or negative. 
Here, an image is classified as negative if it does not contain a person, or if a significant portion of the joints of the person is occluded; otherwise, it is classified as positive.
Similarly, we separate the average confidence score at the server, $c_{s,i}$, into $c_{s,i}^{\mathrm{p}}$ and $c_{s,i}^{\mathrm{n}}$, depending on whether the ground truth of the received image is positive or negative. 
Moreover, we denote the random variables corresponding to $c_{d,i}^{\mathrm{p}}$, $c_{d,i}^{\mathrm{n}}$, $c_{s,i}^{\mathrm{p}}$, and $c_{s,i}^{\mathrm{n}}$ as $C_{d,i}^{\mathrm{p}}$, $C_{d,i}^{\mathrm{n}}$, $C_{s,i}^{\mathrm{p}}$, and $C_{s,i}^{\mathrm{n}}$, respectively, and their cumulative distribution functions (CDFs) as $F_{C_{d,i}^{\mathrm{p}}}(x)$, $F_{C_{d,i}^{\mathrm{n}}}(x)$, $F_{C_{s,i}^{\mathrm{p}}}(x)$, and $F_{C_{s,i}^{\mathrm{n}}}(x)$, respectively. 
That is, $F_{C_{m,i}^{\mathrm{k}}}(x) \!=\! P(C_{m,i}^{\mathrm{k}} \leq x)$ where $\mathrm{k} \in \{\mathrm{p}, \mathrm{n}\}$, $m \in \{d, s\}$, and $i \in \mathcal{N}$.

\begin{figure}[t]
    \centerline{\includegraphics[width=\linewidth]{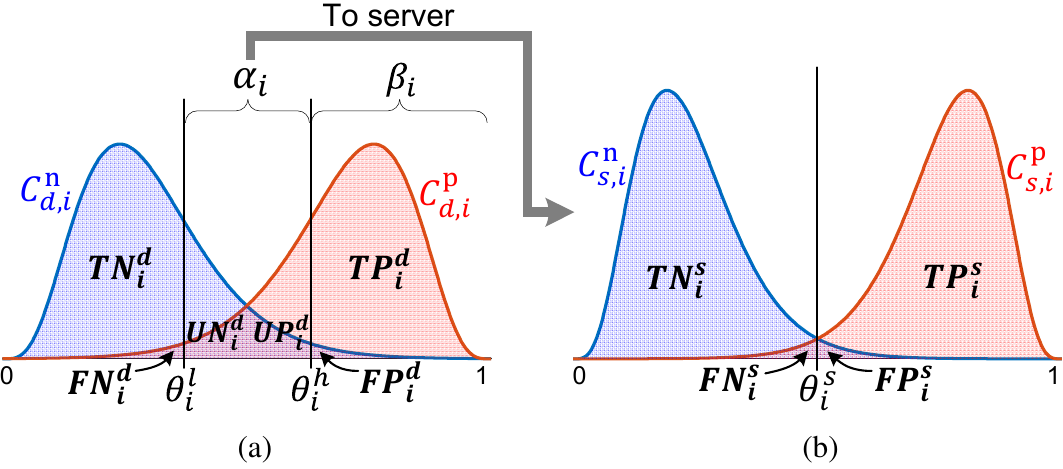}}
    \caption{Confusion matrix analysis for proposed cooperative inference method at (a) end device $i$ and (b) edge server.}
    \label{fig3}
\end{figure}

Fig. \ref{fig3} shows the probability density functions (PDFs) of $C_{d,i}^{\mathrm{p}}$, $C_{d,i}^{\mathrm{n}}$, $C_{s,i}^{\mathrm{p}}$, and $C_{s,i}^{\mathrm{n}}$ for confusion matrix analysis of the proposed cooperative inference method. 
As shown in Fig. \ref{fig2}, device $i$ classifies the input image as \emph{positive} if the confidence score exceeds $\theta_{i}^{h}$. 
Conversely, it classifies the input image as \emph{negative} if the confidence score is below $\theta_{i}^{l}$.
Therefore, as shown in Fig. \ref{fig3}(a), the PDFs of $C_{d,i}^{\mathrm{n}}$ and $C_{d,i}^{\mathrm{p}}$ are partitioned into different regions based on the confidence thresholds $\theta_{i}^{l}$ and $\theta_{i}^{h}$.
Depending on the proportions of these regions, the confusion matrix at device $i$, consisting of true positive (TP), false negative (FN), false positive (FP), and true negative (TN), is expressed as
\begin{align}
T\!P_{i}^{d} &= 1-F_{C_{d,i}^{\mathrm{p}}}(\theta_{i}^{h}), ~~ F\!N_{i}^{d} = F_{C_{d,i}^{\mathrm{p}}}(\theta_{i}^{l}), \label{eq-TPd}\\
F\!P_{i}^{d} &= 1-F_{C_{d,i}^{\mathrm{n}}}(\theta_{i}^{h}), ~~ T\!N_{i}^{d} = F_{C_{d,i}^{\mathrm{n}}}(\theta_{i}^{l}). \label{eq-FPd}
\end{align}

Meanwhile, if the confidence score falls between $\theta_{i}^{l}$ and $\theta_{i}^{h}$, device $i$ classifies the input image as \emph{uncertain}. 
In this context, we define an \emph{uncertain positive (UP)} as a case in which an image is originally positive but has a confidence score that is too ambiguous to be classified as positive. 
Similarly, we define an \emph{uncertain negative (UN)} as a case in which an image is originally negative but has a confidence score that is too ambiguous to be classified as negative.
Accordingly, the UP and UN for device $i$ are respectively expressed as
\begin{align}
U\!P_{i}^{d} &= F_{C_{d,i}^{\mathrm{p}}}(\theta_{i}^{h}) - F_{C_{d,i}^{\mathrm{p}}}(\theta_{i}^{l}), \label{eq-UPd}\\
U\!N_{i}^{d} &= F_{C_{d,i}^{\mathrm{n}}}(\theta_{i}^{h})- F_{C_{d,i}^{\mathrm{n}}}(\theta_{i}^{l}). \label{eq-UNd}
\end{align}

If the server receives images classified as uncertain from device $i$, it re-evaluates them using the confidence threshold $\theta_{i}^{s}$. 
As shown in Fig. \ref{fig3}(b), the PDFs of $C_{s,i}^{\mathrm{p}}$ and $C_{s,i}^{\mathrm{n}}$ are divided into TP, FN, FP, and TN based on $\theta_{i}^{s}$.
Therefore, the confusion matrix at the server for device $i$ is represented as
\begin{align}
T\!P_{i}^{s} &= 1-F_{C_{s,i}^{\mathrm{p}}}(\theta_{i}^{s}), ~~ F\!N_{i}^{s} = F_{C_{s,i}^{\mathrm{p}}}(\theta_{i}^{s}), \label{eq-TPs} \\
F\!P_{i}^{s} &= 1-F_{C_{s,i}^{\mathrm{n}}}(\theta_{i}^{s}), ~~ T\!N_{i}^{s} = F_{C_{s,i}^{\mathrm{n}}}(\theta_{i}^{s}). \label{eq-FPs} 
\end{align}

The final classification at the server is correct if the server classifies the images corresponding to $U\!P_{i}^{d}$ as positive and those corresponding to $U\!N_{i}^{d}$ as negative.
Therefore, the inference accuracy of device $i$ in the proposed cooperative inference method is given by
\begin{align}
A_{i}^{c} &= \frac{T\!P_{i}^{d} + T\!N_{i}^{d} + U\!P_{i}^{d} T\!P_{i}^{s} + U\!N_{i}^{d} T\!N_{i}^{s}}{T\!P_{i}^{d} + F\!N_{i}^{d} + F\!P_{i}^{d} + T\!N_{i}^{d} + U\!P_{i}^{d} + U\!N_{i}^{d}} \nonumber \\ &= \frac{T\!P_{i}^{d} + T\!N_{i}^{d} + U\!P_{i}^{d} T\!P_{i}^{s} + U\!N_{i}^{d} T\!N_{i}^{s}}{2}. \label{eq_Aic}
\end{align}

\subsection{Analysis of End-to-End Delay} \label{subsec:e2eDelay}
The end-to-end delay is defined as the total time elapsed from the moment the images are captured at the devices to when the estimated 3D joint coordinate message arrives at the client.
For simplicity, it is assumed that the images are captured simultaneously across all devices in synchronization with the frame rate and that all devices are homogeneous with the same processing speed.
Therefore, the end-to-end delay is determined by summing the individual delay components, including processing delays at the devices and server, as well as transmission delays from the devices to the client, as follows:
\begin{align}
D_{c} = \underbrace{ T_{d}^{\mathrm{pr}} + T_{d}^{\mathrm{inf}} }_{\mathrm{Delay~at~devices}} + T_{db}^{\mathrm{tx}} +  T_{bs}^{\mathrm{tx}} + \underbrace{ T_{s}^{\mathrm{inf}} + T_{s}^{\mathrm{pr}} }_{\mathrm{Delay~at~server}} + T_{sc}^{\mathrm{tx}}, \label{eq_Dc}
\end{align}
where $T_{d}^{\mathrm{pr}}$ represents the time required for all processing tasks at each device (excluding inference), such as image capture, preprocessing, and transmission preparation, and is treated as a small constant.
Moreover, $T_{d}^{\mathrm{inf}}$ denotes the time required for the device to execute a single inference using its NN model, and can be modeled as $T_{d}^{\mathrm{inf}} = \frac{N_{d} \mathcal{X}_{d}}{\nu_{d}}$, where $N_{d}$ is the number of floating point operations (FLOPs) in the NN of the device, $\mathcal{X}_{d}$ is the number of processor cycles per FLOP on the device, and $\nu_{d}$ is the processor frequency of the device, measured in hertz (Hz) \cite{Choi24ICOIN,Kim2024Distributed}.
Similarly, $T_{s}^{\mathrm{inf}}$ denotes the time required for the server to execute the inferences requested by the devices using its NN model, and the time for a single inference is given by $\frac{N_{s} \mathcal{X}_{s}}{\nu_{s}}$, where $N_{s}$ is the number of FLOPs in the NN of the server, $\mathcal{X}_{s}$ is the number of processor cycles per FLOP on the server, and $\nu_{s}$ is the processor frequency of the server. 
Additionally, $T_{s}^{\mathrm{pr}}$ refers to the processing time required at the server to execute the triangulation algorithm and generate the transmission message, which is also considered a small constant.
Furthermore, $T_{db}^{\mathrm{tx}}$ represents the transmission time required to send data from the devices to the BS for a single 3D pose estimation, $T_{bs}^{\mathrm{tx}}$ denotes the transmission time required to forward data from the BS to the server, and $T_{sc}^{\mathrm{tx}}$ represents the transmission time required to send the 3D joint coordinate message from the server to the client.


As shown in Fig. \ref{fig3}(a), we denote $\alpha_{i}$ as the probability that device $i$ classifies an input image as uncertain, which is given by
\begin{align}
\alpha_{i} &= \frac{U\!P_{i}^{d} + U\!N_{i}^{d}}{2}. \label{eq_beta}
\end{align}
As each device requests re-inference to the server with a probability of $\alpha_{i}$, the total average $T_{s}^{\mathrm{inf}}$ is calculated as
\begin{align}
\bar{T}_{s}^{\mathrm{inf}} = \sum_{i=1}^{N} \alpha_{i} \frac{N_{s} \mathcal{X}_{s}}{\nu_{s}}. \label{eq_Tsinf}
\end{align}


In addition, we denote $\beta_{i}$ as the probability that device $i$ classifies an input image as positive, which is given by
\begin{align}
\beta_{i} &= \frac{T\!P_{i}^{d} + F\!P_{i}^{d}}{2}. \label{eq_alpha}
\end{align}
As device $i$ transmits an image to the BS with probability $\alpha_{i}$ and a message with probability $\beta_{i}$, the average data size transmitted from device $i$ to the BS per second is given by
\begin{align}
B_{i} = f (\alpha_{i}S_{g} + \beta_{i} S_{m}),
\end{align}
where $f$ is the number of images input to the devices per second, and $S_{g}$ and $S_{m}$ are the sizes of an image and a message in bits, respectively. 
Therefore, the average $T_{db}^{\mathrm{tx}}$ is calculated as
\begin{align}
\bar{T}_{db}^{\mathrm{tx}} = \sum_{i=1}^{N} \frac{B_{i}}{R_{i}} = \sum_{i=1}^{N} \frac{f (\alpha_{i} S_{g} + \beta_{i} S_{m})}{R_{i}}, \label{eq_Tdbtx}
\end{align}
where $R_{i}$ is the wireless transmission rate from device $i$ to the BS.
From the Shannon theorem, $R_{i}$ is given by 
\begin{align}
R_{i} = \tau_{i} W \log_{2} \left( 1+\frac{P g_{i}}{N_{0}W} \right),
\end{align}
where $\tau_{i}$ denotes the time allocation ratio for device $i$, subject to $\sum_{i=1}^{N} \tau_{i} \leq 1$, $W$ is the bandwidth of the wireless channel, $P$ is the constant transmission power of the devices, and $N_{0}$ is the power spectral density of the noise.

As the BS directly forwards the data received from the devices to the server via a wired link, the average $T_{bs}^{\mathrm{tx}}$ is calculated as
\begin{align}
\bar{T}_{bs}^{\mathrm{tx}} = \frac{\sum_{i=1}^{N} B_{i}}{R_{b}} = \frac{\sum_{i=1}^{N} f (\alpha_{i} S_{g} + \beta_{i} S_{m})}{R_{b}}, \label{eq_Tbstx}
\end{align}
where $R_{b}$ represents the constant transmission rate of the wired backhaul link between the BS and the server.
Finally, as the server transmits the 3D joint coordinate message to the client, $T_{sc}^{\mathrm{tx}}$ is given by
\begin{align}
T_{sc}^{\mathrm{tx}} = \frac{S_{c}}{R_{c}}, \label{eq_Tsdtx}
\end{align}
where $S_{c}$ is the size of the 3D joint coordinate message, and $R_{c}$ is the constant transmission rate between the server and client, assuming that channel variations in wired links are negligible \cite{olshefski2002inferring}.

Consequently, by applying the derived delay components \eqref{eq_Tsinf}, \eqref{eq_Tdbtx}, \eqref{eq_Tbstx}, and \eqref{eq_Tsdtx} to \eqref{eq_Dc}, the average end-to-end delay of the proposed cooperative inference method is expressed as
\begin{align}
\bar{D}_{c} &= T_{d}^{\mathrm{pr}} + T_{d}^{\mathrm{inf}} + \bar{T}_{db}^{\mathrm{tx}} +  \bar{T}_{bs}^{\mathrm{tx}} + \bar{T}_{s}^{\mathrm{inf}} + T_{s}^{\mathrm{pr}} + T_{sc}^{\mathrm{tx}} \nonumber \\
&= T_{d}^{\mathrm{pr}} + \frac{N_{d} \mathcal{X}_{d}}{\nu_{d}} + \sum_{i=1}^{N} \frac{f (\alpha_{i} S_{g} + \beta_{i} S_{m})}{R_{i}} \nonumber \\ & ~~~ + \sum_{i=1}^{N} \frac{f (\alpha_{i} S_{g} + \beta_{i} S_{m})}{R_{b}} + \sum_{i=1}^{N} \alpha_{i} \frac{N_{s} \mathcal{X}_{s}}{\nu_{s}} + T_{s}^{\mathrm{pr}} + \frac{S_{c}}{R_{c}}.
\label{eq_aveDc}
\end{align}

\subsection{Analysis of Conventional Inference Methods}
The proposed cooperative inference method can be considered as a generalized framework that encompasses conventional inference methods by setting the two thresholds of the end devices to specific values.
First, if the threshold values are set as $\theta_{i}^{l} = \theta_{i}^{h}$, $\forall i$, the operation of the proposed cooperative inference method becomes equivalent to that of the device-centric inference method because all 2D pose estimations are performed only on the end devices \cite{choi2021mobilehumanpose}.
This threshold setting results in $U\!P_{i}^{d} = U\!N_{i}^{d} = 0$ and $\alpha_{i} = 0$, $\forall i$.
Applying these values to \eqref{eq_Aic}, the inference accuracy of device $i$ in the device-centric inference method is obtained as
\begin{align}
A_{i}^{d} = \frac{T\!P_{i}^{d} + T\!N_{i}^{d}}{2}.
\end{align}
Moreover, by applying these values to \eqref{eq_aveDc}, the average end-to-end delay of the device-centric inference method is obtained as 
\begin{align}
\bar{D}_{d} \!=\! T_{d}^{\mathrm{pr}} \!+\! \frac{N_{d} \mathcal{X}_{d}}{\nu_{d}} \!+\! \sum_{i=1}^{N} \frac{f \beta_{i} S_{m}}{R_{i}} \!+\! \sum_{i=1}^{N} \frac{f \beta_{i} S_{m}}{R_{b}} \!+\! T_{s}^{\mathrm{pr}} \!+\! \frac{S_{c}}{R_{c}}.
\end{align}

Second, if the threshold values are set as $\theta_{i}^{l} = 0$ and $\theta_{i}^{h} = 1$, $\forall i$, the operation of the proposed cooperative inference method becomes equivalent to that of the server-centric inference method because all 2D pose estimations are performed only on the edge server \cite{chang2021pose}.
This threshold setting results in $T\!P_{i}^{d} = F\!N_{i}^{d} = F\!P_{i}^{d} = T\!N_{i}^{d} = 0$, $U\!P_{i}^{d} = U\!N_{i}^{d} = 1$, $\alpha_{i} = 1$, and $\beta_{i} = 0$, $\forall i$, and also allows for $T_{d}^{\mathrm{inf}} = 0$. 
By applying these values to \eqref{eq_Aic}, the inference accuracy of device $i$ in the server-centric inference method is obtained as
\begin{align}
A_{i}^{s} = \frac{T\!P_{i}^{s} + T\!N_{i}^{s}}{2}.
\end{align}
Similarly, by applying these values to \eqref{eq_aveDc}, the average end-to-end delay of the server-centric inference method is obtained as 
\begin{align}
\bar{D}_{s} = T_{d}^{\mathrm{pr}} + \sum_{i=1}^{N}\frac{f S_{g}}{ R_{i}} + \frac{N f S_{g}}{R_{b}} + \frac{N N_{s} \mathcal{X}_{s}}{\nu_{s}} + T_{s}^{\mathrm{pr}} + \frac{S_{c}}{R_{c}}.
\end{align}


Third, the operation of the proposed cooperative inference method aligns with that of the conventional cascade inference method when $\theta_{i}^{l}$ is fixed at zero and only $\theta_{i}^{h}$ is utilized, as cascade inference employs a single confidence threshold at the device \cite{zhao2019deep,enomoro2021learning}.
In cascade inference, if the confidence score is greater than $\theta_{i}^{h}$, the device transmits its inference result to the server, whereas if it is lower, the device requests re-inference to the server. 
As a result, $U\!N_{i}^{d} = T\!N_{i}^{d}$ and $U\!P_{i}^{d} = F\!N_{i}^{d}$ are established.  
Applying these to \eqref{eq_Aic}, the inference accuracy of device $i$ in the cascade inference method is expressed as
\begin{align}
A_{i}^{a} = \!\! \frac{T\!P_{i}^{d} + T\!N_{i}^{d} + F\!N_{i}^{d} T\!P_{i}^{s} + T\!N_{i}^{d} T\!N_{i}^{s}}{2}. \label{eq_Aia}
\end{align}
Additionally, in this case, $\beta_{i} = 1-\alpha_{i}$ holds. 
By applying this to \eqref{eq_aveDc}, the average end-to-end delay of the cascade inference method is expressed as
\begin{align}
\bar{D}_{a} &= T_{d}^{\mathrm{pr}} + \frac{N_{d} \mathcal{X}_{d}}{\nu_{d}} + \sum_{i=1}^{N} \frac{f (\alpha_{i} S_{g} + (1-\alpha_{i}) S_{m})}{R_{i}} \nonumber \\ & ~~ + \sum_{i=1}^{N} \frac{f (\alpha_{i} S_{g} \!+\! (1\!-\!\alpha_{i}) S_{m})}{R_{b}} \!+\! \sum_{i=1}^{N} \alpha_{i} \frac{N_{s} \mathcal{X}_{s}}{\nu_{s}} \!+\! T_{s}^{\mathrm{pr}} \!+\! \frac{S_{c}}{R_{c}}.
\label{eq_aveDa}
\end{align}

\section{Joint Optimization}
\label{title:opt}
The performance of 3D human pose estimation is typically assessed using the MPJPE, which is defined as {\cite{bultmann2021real}}
\begin{align}
\mbox{MPJPE} \triangleq \frac{1}{J} \sum_{j=1}^{J} \big|\big| \mathbf{x}^{j} - \hat{\mathbf{x}}^{j} \big|\big|,
\end{align}
where $\mathbf{x}^{j}$ and $\hat{\mathbf{x}}^{j}$ represent the ground truth and estimated 3D coordinates of joint $j \!\in\! \mathcal{J}$, respectively.
Our objective is to achieve accurate and real-time 3D human pose estimation using the proposed cooperative inference method.
Therefore, we formulate a joint optimization problem to derive the optimal confidence thresholds and transmission time for each device, aiming to minimize the MPJPE while satisfying the end-to-end delay requirement, as follows:
\begin{align}
\textbf{(P1):} ~\min_{\pmb{\theta}_{l}, \pmb{\theta}_{h}, \pmb{\theta}_{s}, \pmb{\tau}} &~ \frac{1}{J} \sum_{j=1}^{J} \big|\big| \mathbf{x}^{j} - \hat{\mathbf{x}}^{j} \big|\big| \\
\mbox{s.t.} ~~~ &~ \bar{D}_{c} \leq D_{req}, \label{const1} \\
&~0 \leq \theta_{i}^{l} \leq \theta_{i}^{h} \leq 1, ~ 0 \leq \theta_{i}^{s} \leq 1, ~\forall i \!\in\! \mathcal{N}, \label{const2} \\
&~\textstyle \sum_{i=1}^{N} \tau_{i} \leq 1, ~ 0 \leq \tau_{i} \leq 1, ~\forall i \!\in\! \mathcal{N}, \label{const3} 
\end{align}
where the optimization variables are defined as $\pmb{\theta}_{l} \triangleq \{\theta_{i}^{l}, ~\forall i\}$, $\pmb{\theta}_{h} \triangleq \{\theta_{i}^{h}, ~\forall i\}$, $\pmb{\theta}_{s} \triangleq \{\theta_{i}^{s}, ~\forall i\}$, and $\pmb{\tau} \triangleq \{\tau_i, ~\forall i\}$. 
Further, $D_{req}$ denotes the required end-to-end delay constraint.

Since solving \textbf{(P1)} with the MPJPE as the objective function is mathematically nontrivial, we introduce the following Lemma.
\begin{lemma} \label{lemma1}
Minimizing the MPJPE is equivalent to maximizing the sum of the inference accuracies for all devices.
That is,
\begin{align}
\min_{\pmb{\theta}} \frac{1}{J} \sum_{j=1}^{J} \big|\big| \mathbf{x}^{j} - \hat{\mathbf{x}}^{j} \big|\big|
\Longleftrightarrow
\max_{\pmb{\theta}} \sum_{i=1}^{N} A_{i}, 
\label{eq_lemma}
\end{align}
where $\pmb{\theta}\triangleq\{\pmb{\theta}_{l}, \pmb{\theta}_{h}, \pmb{\theta}_{s}\}$ is defined.
\end{lemma}
\begin{IEEEproof}
We can rewrite $\hat{\mathbf{x}}^{j} = \mathbf{e}^{j} \circ \mathbf{x}^{j}$ where $\mathbf{e}^{j} = [e_{x}^{j}, e_{y}^{j}, e_{z}^{j}] \in \mathbb{R}^{3}$ represents the estimation deviation vector for joint $j$, and $\circ$ denotes element-wise product. 
Thus, the expression $\big|\big| \mathbf{x}^{j} - \hat{\mathbf{x}}^{j} \big|\big|$ can be written as
\begin{align}
\big|\big| \mathbf{x}^{j} - \hat{\mathbf{x}}^{j} \big|\big| = \big|\big| (\mathbf{1}-\mathbf{e}^{j}) \circ \mathbf{x}^{j} \big|\big|.
\end{align}
Intuitively, as the confidence scores for positive images increase and those for negative images decrease, $\mathbf{e}^{j}$ tends to approach the ones vector $\mathbf{1}_{(3)}$.
That is, if the confidence scores for joint $j$ detected from positive images fed into device $i$ and used for the MPJPE calculation, denoted as $c_{i}^{j,\mathrm{p}}$, approach one, and the confidence scores for joint $j$ detected from negative images fed into device $i$ and used for the MPJPE calculation, denoted as $c_{i}^{j,\mathrm{n}}$, approach zero, then $\mathbf{e}^{j}$ converges to $\mathbf{1}_{(3)}$.\footnote{Note that the subscript $d$ or $s$ is omitted in $c_{i}^{j,\mathrm{p}}$ and $c_{i}^{j,\mathrm{n}}$, as these values are determined by the device, server, or their cooperation, depending on the inference method.}
Therefore, as $c_{i}^{j,\mathrm{p}} \rightarrow 1$ and $c_{i}^{j,\mathrm{n}} \rightarrow 0$, the term $\big|\big| (\mathbf{1}-\mathbf{e}^{j}) \circ \mathbf{x}^{j} \big|\big| \rightarrow 0$ and $\big|\big| \mathbf{x}^{j} - \hat{\mathbf{x}}^{j} \big|\big| \rightarrow 0$, which leads to a reduction in the MPJPE.
Accordingly, the left-hand side of \eqref{eq_lemma} can be rewritten as follows:
\begin{align}
\min_{\pmb{\theta}} \frac{1}{J} \sum_{j=1}^{J} \big|\big| \mathbf{x}^{j} - \hat{\mathbf{x}}^{j} \big|\big|
\Longleftrightarrow
\max_{\pmb{\theta}} \frac{1}{J} \sum_{j=1}^{J} \left( \sum_{i=1}^{N} \!c_{i}^{j,\mathrm{p}} \!-\! \sum_{i=1}^{N} \!c_{i}^{j,\mathrm{n}} \right). \label{eq_maxC}
\end{align}
By representing the average values as $\frac{1}{J} \sum_{j=1}^{J} c_{i}^{j,\mathrm{p}} = c_{i}^{\mathrm{p}}$ and $\frac{1}{J} \sum_{j=1}^{J} c_{i}^{j,\mathrm{n}} = c_{i}^{\mathrm{n}}$, the right-hand side of \eqref{eq_maxC} is expressed as 
\begin{align}
\frac{1}{J} \sum_{j=1}^{J} \left( \sum_{i=1}^{N} \!c_{i}^{j,\mathrm{p}} \!-\! \sum_{i=1}^{N} \!c_{i}^{j,\mathrm{n}} \right) & = \sum_{i=1}^{N} \left( c_{i}^{\mathrm{p}} - c_{i}^{\mathrm{n}} \right) \nonumber \\
& \hspace{-2.3cm} = \sum_{i=1}^{N} \left( \int_{\tilde{\theta}_{i}}^{1} \!x f_{C_{i}^{\mathrm{p}}}(x) dx -\! \int_{\tilde{\theta}_{i}}^{1} \!x f_{C_{i}^{\mathrm{n}}}(x) dx \right), 
\label{eq_XpXn}
\end{align}
where $\tilde{\theta}_{i}$ serves as an effective confidence threshold for filtering images used in the MPJPE calculation, determined based on $\theta_{i}^{l}$, $\theta_{i}^{h}$, and $\theta_{i}^{s}$.
By applying \eqref{eq_XpXn} to \eqref{eq_maxC}, we can build the following relationship:
\begin{align}
\min_{\pmb{\theta}} & \frac{1}{J} \sum_{j=1}^{J} \big|\big| \mathbf{x}^{j} - \hat{\mathbf{x}}^{j} \big|\big| \nonumber \\ & \Longleftrightarrow \max_{\pmb{\theta}} \sum_{i=1}^{N} \!\left( \int_{\tilde{\theta}_{i}}^{1} \!x f_{C_{i}^{\mathrm{p}}}(x) dx \!-\! \int_{\tilde{\theta}_{i}}^{1} \!x f_{C_{i}^{\mathrm{n}}}(x) dx \right) \nonumber \\ 
& \Longleftrightarrow 
\max_{\pmb{\theta}} \sum_{i=1}^{N} \!\left( \int_{\tilde{\theta}_{i}}^{1} \! f_{C_{i}^{\mathrm{p}}}(x) dx \!-\! \int_{\tilde{\theta}_{i}}^{1} \! f_{C_{i}^{\mathrm{n}}}(x) dx \right) \nonumber \\ 
&~~~~~~ = \max_{\pmb{\theta}} \sum_{i=1}^{N} \left( TP_{i} - FP_{i} \right) \nonumber \\
&~~~~~~ = \max_{\pmb{\theta}} \sum_{i=1}^{N} \left( TP_{i} + TN_{i} - 1 \right)  \nonumber \\
& \Longleftrightarrow 
\max_{\pmb{\theta}} \sum_{i=1}^{N} \frac{TP_{i} + TN_{i}}{2} = \max_{\pmb{\theta}} \sum_{i=1}^{N} A_{i}.
\end{align}
\end{IEEEproof}

From Lemma \ref{lemma1}, we can convert the original problem \textbf{(P1)} into an optimization problem that maximizes the sum of the inference accuracies for all the devices, as follows:
\begin{align}
\textbf{(P2):} ~\max_{\pmb{\theta}, \pmb{\tau}} &~~ \sum_{i=1}^{N} A_{i}^{c}, \nonumber \\
\mbox{s.t.}~ &~~ \eqref{const1},~ \eqref{const2},~ \eqref{const3}. \nonumber
\end{align}
Considering that $A_{i}^{c}$ and $\bar{D}_{c}$ cannot be expressed as explicit equations in terms of $\pmb{\theta}$, and \textbf{(P2)} is not jointly concave with respect to $\pmb{\theta}$ and $\pmb{\tau}$, it is analytically intractable to solve \textbf{(P2)} directly.
Therefore, we decompose problem \textbf{(P2)} into two subproblems, each focusing on one of the optimization variables, $\pmb{\theta}$ or $\pmb{\tau}$. 
We then iteratively solve each subproblem by fixing one variable while optimizing the other to find suboptimal solutions.

First, when $\pmb{\theta}$ is fixed, $A_{i}^{c}$ is not explicitly a function of $\pmb{\tau}$, but $\bar{D}_{c}$ is directly affected by $\pmb{\tau}$, as described in Subsection \ref{subsec:e2eDelay}.
As the NN model on the server provides superior accuracy performance compared to that of the devices, it is necessary for the devices to offload more images to the server to enhance the accuracy.
To achieve this, we first determine $\pmb{\tau}$ that minimizes $\bar{D}_{c}$, and then adjust $\pmb{\theta}$ to enable more offloading to the server while satisfying $\bar{D}_{c} \leq D_{req}$ under the obtained $\pmb{\tau}$. 
Therefore, we formulate the first subproblem as follows:
\begin{align}
\textbf{(P2-1):} ~\min_{\pmb{\tau}} ~&~ \bar{D}_{c}, \nonumber \\
\mbox{s.t.}~~ &~ \eqref{const1},~ \eqref{const3}. \nonumber
\end{align}
Since \textbf{(P2-1)} is a convex problem and the strong duality holds, we can find the optimal value of $\pmb{\tau}$ using a dual method.
First, the Lagrangian function of \textbf{(P2-1)} is defined as
\begin{align}\label{Lag}
\mathcal{L}(\pmb{\tau}, \lambda, \mu) = \bar{D}_{c} + \lambda \left( \bar{D}_{c}- D_{req} \right) + \mu \left( \sum_{i=1}^{N} \tau_{i}- 1 \right),
\end{align}
where $\lambda \geq 0$ and $\mu \geq 0$ are the Lagrange multipliers for each constraint in \textbf{(P2-1)}. 
Therefore, the dual function is defined as 
\begin{align}\label{df}
\mathcal{G}(\lambda, \mu) = \min_{0 \preceq \pmb{\tau} \preceq 1} \mathcal{L}(\pmb{\tau}, \lambda, \mu).
\end{align}
The dual problem is formulated as 
\begin{align}
\max_{\lambda \geq 0, ~ \mu \geq 0} \mathcal{G}(\lambda, \mu).
\label{Dual2}
\end{align}
From the Karush-Kuhn-Tucker (KKT) condition $\frac{\partial \mathcal{L}(\pmb{\tau}, \lambda, \mu)}{\partial \tau_i} = 0$, the optimal value of $\tau_i$ that minimizes $\mathcal{L}(\pmb{\tau}, \lambda, \mu)$ is derived as  
\begin{align}
\tau_i = \left[\sqrt{\frac{(1+\lambda)f (\alpha_{i} S_{g} + \beta_{i} S_{m})}{\mu W \log_{2} \left( 1+\frac{P g_{i}}{N_{0}W} \right)}}\right]_0^1,
\label{opt_tau}
\end{align}
where $[\cdot]_0^1=\max(\min(\cdot,1),0)$.
Subsequently, the Lagrange multipliers are optimally updated using a gradient algorithm because the dual problem is convex with respect to $\lambda$ and $\mu$, as follows:
\begin{align}
\displaystyle  \lambda &\leftarrow \left[ \lambda + \kappa_1 \left( \bar{D}_{c}- D_{req} \right)  \right]^+, \label{LM1}\\
\mu &\leftarrow \left[ \mu + \kappa_2 \left( \sum_{i=1}^{N} \tau_{i}- 1 \right)  \right]^+, \label{LM2}
\end{align}
where $\kappa_1$ and $\kappa_2$ are the step sizes for the update and are chosen to be sufficiently small. 
Then, the optimal $\pmb{\tau}$ is determined by iteratively updating $\tau_i$ according to \eqref{opt_tau} and $(\lambda,\mu)$ according to \eqref{LM1} and \eqref{LM2} until convergence is achieved. 

Next, when $\pmb{\tau}$ is fixed, we can formulate the subproblem for $\pmb{\theta}$ as follows:
\begin{align}
\textbf{(P2-2):} ~\max_{\pmb{\theta}} ~& ~ \sum_{i=1}^{N} A_{i}^{c}, \nonumber \\
\mbox{s.t.}~~ &~ \eqref{const1},~ \eqref{const2}. \nonumber
\end{align}
As $A_{i}^{c}$ cannot be explicitly expressed in terms of $\pmb{\theta}$, analytical optimization methods cannot be directly applied to solve \textbf{(P2-2)}. 
Therefore, a greedy search method is used to solve the problem efficiently and with low complexity.
Specifically, the threshold $\pmb{\theta}_{s}$ of the edge server was first fixed and the two thresholds $\pmb{\theta}_{l}$ and $\pmb{\theta}_{h}$ of the end devices that maximize $\sum_{i=1}^{N} A_{i}^{c}$ are determined. 
With the optimized thresholds $\pmb{\theta}_{l}^{*}$ and $\pmb{\theta}_{h}^{*}$ obtained, the optimal threshold $\pmb{\theta}_{s}^{*}$ that maximize $\sum_{i=1}^{N} A_{i}^{c}$ is then determined. 
This process is repeated until all thresholds converge. 

It is worth noting that in \textbf{(P2-1)}, the objective is to minimize the delay, which yields a solution in which the delay constraint \eqref{const1} is satisfied with some margin.
Subsequently, \textbf{(P2-2)} optimizes $\pmb{\theta}$ within the expanded feasible range of constraint \eqref{const1}, allowing more images to be offloaded to the server at the expense of a slight increase in delay, which in turn leads to enhanced accuracy.
In other words, the minimized delay from \textbf{(P2-1)} effectively expands the feasible search space of $\pmb{\theta}$, providing more flexibility to enhance accuracy without violating the delay constraint.
Therefore, by iteratively solving \textbf{(P2-1)} and \textbf{(P2-2)}, $\pmb{\theta}$ and $\pmb{\tau}$ are jointly optimized to maximize accuracy while tightly satisfying the delay constraint \eqref{const1}, thereby achieving a near-optimal performance, as demonstrated in the simulation results (e.g., Fig. \ref{graph4}).

Algorithm \ref{Alg1} summarizes the overall procedure for finding the optimal $\pmb{\theta}$ and $\pmb{\tau}$. 
Initially, the optimization variables are set to feasible values. 
The algorithm then optimizes $\pmb{\tau}$ by solving $\textbf{(P2-1)}$ for the given $\pmb{\theta}$, as shown in Lines 5--8.
Thereafter, the algorithm continues to optimize $\pmb{\theta}$ under the determined $\pmb{\tau}$ based on the greedy search method, as shown in Lines 10--23.
Specifically, given $\pmb{\tau}[t]$, it finds the optimal $\theta_{i}^{l}, \theta_{i}^{h}, \theta_{i}^{s}$ and updates these values to $\pmb{\theta}[t]$ for the next iteration.
These operations are repeated sequentially until all optimization variables converge.

\begin{algorithm}[h]
\caption{Proposed Optimization Algorithm} \label{Alg1} \small
\begin{algorithmic}[1]
\State Initialize $t \!=\! 0$, $\pmb{\theta}[t]\!\triangleq\!\{ \theta^l_{i}[t], \theta^h_{i}[t], \theta_{i}^{s}[t], \forall i\}\!=\! 0.5$, $\pmb{\tau}[t]\!\triangleq\!\{\tau_i[t],\forall i\}\!=\! \frac{1}{N}$
\Repeat
    \State $t \leftarrow t+1$
    \State /* Find the optimal $\pmb{\tau}$ */
    \Repeat
       \State Update $\pmb{\tau}[t]$ according to \eqref{opt_tau} 
       \State Update $\lambda$ and $\mu$ according to \eqref{LM1} and \eqref{LM2}
    \Until {Convergence of $\pmb{\tau}[t]$}
    \vspace{1mm}
    \State /* Find the optimal $\pmb{\theta}_{l}$, $\pmb{\theta}_{h}$, and $\pmb{\theta}_{s}$ */
    \State Initialize $k\!=\!0$ and $\theta_{i}^{s}[k] \!=\! \theta_{i}^{s}[t\!-\!1]$, $\forall i$
    \Repeat
        \State $k \leftarrow k+1$
        \State Update $\theta_{i}^{s} \leftarrow \theta_{i}^{s}[k\!-\!1]$, $\forall i$
        \For {$\theta^{l}_i, \theta^{h}_i = [0, 1]$}      
            \If {$\theta^{l}_i \leq \theta^{h}_i$}
            \State Find $\theta^{l}_i[k]$ and $\theta^{h}_i[k]$, $\forall i$, by solving \textbf{(P2-2)}
            \EndIf
        \EndFor
        \State Update $\theta^{l}_i \leftarrow \theta^{l}_i[k]$ and $\theta^{h}_i \leftarrow \theta^{h}_i[k]$, $\forall i$
        \For {$\theta_{i}^{s} = [0, 1]$}
            \State Find $\theta_{i}^{s}[k]$, $\forall i$, by solving \textbf{(P2-2)} 
        \EndFor          
    \Until {Convergence of $\{\theta^{l}_i[k], \theta^{h}_i[k], \theta_{i}^{s}[k]\}$}
    \State Update $\pmb{\theta}[t] \leftarrow \pmb{\theta}[k]$
    
\Until Convergence of all variables
\end{algorithmic}
\end{algorithm}

Furthermore, the convergence and complexity of the proposed optimization algorithm are analyzed in the following two remarks.

\begin{remark}[Proof of Convergence] \label{remark2}
The convergence of Algorithm 1 can be proved as follows. 
Let the function be defined as $f(\pmb{\theta}, \pmb{\tau}) = \sum_{i=1}^{N} A_{i}^{c}$.
At the $t$-th iteration, the optimal solution of $\pmb{\tau}$, denoted as $\pmb{\tau}[t]$, can be found for a given $\pmb{\theta}[t-1]$.
Thus, the following equation holds for any $\pmb{\tau}[t-1]$. 
\begin{align}
f \left(\pmb{\theta}[t-1], \pmb{\tau}[t-1] \right) = f \left(\pmb{\theta}[t-1], \pmb{\tau}[t] \right), \label{convergence1}
\end{align}
where the equality holds because the update of $\pmb{\tau}$ does not affect the accuracy improvements.
Moreover, at the $t$-th iteration, the optimal solution of $\pmb{\theta}$, denoted as $\pmb{\theta}[t]$, is found for a given $\pmb{\tau}[t]$, then the following inequality holds for any $\pmb{\theta}[t-1]$.
\begin{align}
f \left(\pmb{\theta}[t-1], \pmb{\tau}[t]\right) \leq f \left(\pmb{\theta}[t], \pmb{\tau}[t]\right). \label{convergence2}
\end{align}
Therefore, from the non-decreasing nature of the objective value in each iteration of Algorithm 1, as shown in \eqref{convergence1} and \eqref{convergence2}, we can conclude that 
\begin{align}
f \left(\pmb{\theta}[t-1], \pmb{\tau}[t-1] \right) \leq f \left(\pmb{\theta}[t], \pmb{\tau}[t] \right).
\end{align}
Furthermore, due to the finite range of the optimization variables, the objective value is upper bounded by a finite value \cite{Bertsekas99}, which guarantees the convergence of Algorithm 1. \qed
\end{remark}

\begin{remark}[Analysis of Complexity]
\label{remark3}
Adopting the methodology to calculate the computational complexity of the worst-case interior point method \cite{Ben-Tal01,Boyd04}, the computational complexity of finding the optimal $\pmb{\tau}$ in Algorithm \ref{Alg1} can be derived as $\mathcal{O}\left(N\log(1/\epsilon)\right)$, where $\epsilon > 0$ is the convergence threshold.
Moreover, the computational complexity of finding the optimal $\pmb{\theta}$ in Algorithm \ref{Alg1} can be expressed as $\mathcal{O}(NKM^{2})$, where $M$ is the number of segments that divide the three thresholds into equal intervals and $K$ is the number of iterations required for the convergence of the inner loop in Lines 11--23. 
Therefore, the overall complexity of Algorithm \ref{Alg1} is given by $\mathcal{O}\left(TN(\log(1/\epsilon)+KM^{2}) \right)$, where $T$ indicates the number of time slots required for the convergence of the outer loop in Lines 2--25.
\end{remark}

\section{Results and Discussion}
\label{title:result}

\begin{figure}[t]
    \centerline{\includegraphics[width=\linewidth]{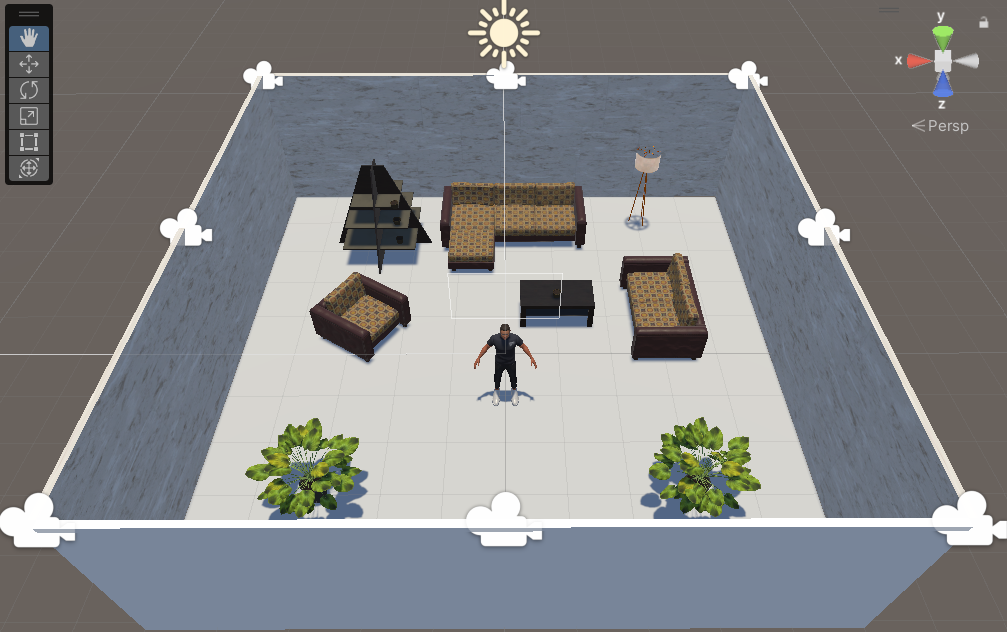}}
    \caption{Virtual experimental environment implemented in Unity.}
    \label{fig4}
\end{figure}

To facilitate the acquisition of ground-truth 3D joint coordinates and the capture of test images from multiple cameras for diverse human poses in motion, we considered a virtual environment implemented in Unity, as shown in Fig. \ref{fig4}.\footnote{The source code and a demonstration video for this implementation are available at \cite{Choi25hkpose3d}.}
As a virtual environment enables multiple cameras to be easily positioned at the desired locations and provides ground-truth data with reduced errors for the 3D joint coordinates of various human poses, it has often been used as a cost-effective alternative to experiments conducted in real-world environments \cite{Qureshi08,Tian18}.
In line with this, this study focused on the validation and optimization of the proposed cooperative inference method, which required large-scale images and ground-truth data.
Therefore, a virtual experimental environment was adopted.\footnote{Future research will verify the performance of the proposed method in real-world environments.}
For data collection, a virtual indoor room was designed within a cubic space of 10$\times$10$\times$3 $m^{3}$. 
Eight cameras were positioned at the corners and along the edges, with their focal points directed toward the center of the room.
A humanoid avatar was placed in this space and programmed to move uniformly at random across all points in the room for 20 min.
During this movement, the avatar performed random actions, such as spreading its arms, falling, and jumping.
All cameras captured images at a rate of 2 frames per second, and simultaneously, the ground-truth information for the 3D joint coordinates of the person was recorded.

\begin{table}[t]
\begin{center}
\caption{Parameter Setup}
\begin{tabular}{ll} \toprule 
Description & Value \\ \midrule
NN model in device & YOLOv8-pose nano \\ 
NN model in server & YOLOv8-pose xlarge \\
Number of joints detected & $J$ = 17 \\
Number of devices & $N$ = 2$\sim$8 (default \!=\! 4) \\
Frame per second & $f$ = 2 fps \\
Size of image transmitted & $S_{g}$ = 2$\sim$128 KB (default \!=\! 32) \\ 
Size of message & $S_{m}$ = 68 bytes \\
Inference time at device & $T_{d}^{\mathrm{inf}}$ = 20$\sim$500 ms (default \!=\! 100) \\
Inference time at server & $T_{s}^{\mathrm{inf}}$ = 20 ms \\ 
End-to-end delay requirement & $D_{req}$ = 0.2$\sim$1 s (default \!=\! 0.5) \\
Processing time at device & $T_{d}^{\mathrm{pr}}$ = 10 ms \\
Processing time at server & $T_{s}^{\mathrm{pr}}$ = 5 ms \\
Tx. time from BS to server & $T_{bs}^{\mathrm{tx}}$ = 0.5 ms \\
Tx. time from server to client & $T_{sc}^{\mathrm{tx}}$ = 20 ms \\ 
Channel bandwidth & $W$ = 1 MHz \\
Noise power spectral density & $N_{0}$ = --165 dBm/Hz \\
Transmission power at device \!\!\!\!\! & $P$ = 30 dBm \\
Average channel gain & $g$ = --130$\sim$--60 dB (default \!=\! --100) \\ 
\bottomrule
\end{tabular}
\label{table_para}
\end{center}
\end{table}

Table \ref{table_para} summarizes the parameters used for performance evaluation. 
The YOLOv8 pose estimation model was used as a state-of-the-art deep learning framework that infers the coordinates of the 17 key human joints \cite{ultralytics_yolov8_docs}.
Among the available YOLOv8 model sizes, the smallest \emph{nano} model was chosen for the end devices, whereas the largest \emph{xlarge} model was selected for the edge server. 
It was assumed that the devices captured images at full HD resolution, which were then compressed into the JPEG format for inference and transmission \cite{rego2018intelligent}. 
The average size of a JPEG image ($S_{g}$) was 32 kilobytes.
In contrast, the message size ($S_{m}$) was set to 68 bytes, assuming that the $(u,v)$ coordinates of the 17 joints were encoded using 2 bytes per coordinate.
It was also assumed that the device utilized an Intel NUC Core i5 CPU, and the server employed an NVIDIA RTX 4090 GPU for inference. 
The range of inference times measured on each processor was determined based on the average values from multiple tests, and the default time for a single inference was set to 100 ms on the device and 20 ms on the server according to the inference times specified in YOLOv8 documentation
\cite{ultralytics_yolov8_docs}.
Furthermore, the transmission times between each node were set to appropriate values, considering the actual network conditions \cite{kekki2018mec,olshefski2002inferring}.
Other parameters related to the wireless channel and power ($W$, $N_{0}$, $g$, and $P$) were selected based on values commonly used in \cite{Mei23Energy,Kim2024Distributed,amanatidis2023cooperative}.
In particular, the average channel gain $g$ varied within a practical range considering path loss and fading effects. The channel gain for each device was assigned according to a normal distribution.
Unless otherwise stated, the default values were used for all results.

\begin{figure}[t]
    \centerline{\includegraphics[width=0.9\linewidth]{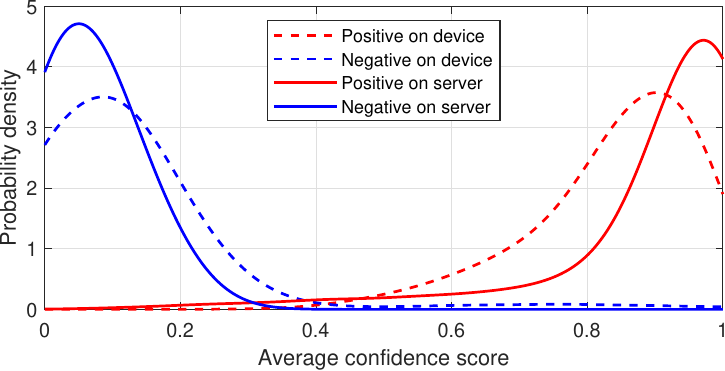}}
    \caption{PDFs of average confidence scores on the device and server.}
    \label{graph1}
\end{figure}

Fig. \ref{graph1} presents the PDFs of the average confidence scores for all the joints from the positive and negative images on both the device and the server (i.e., $C_{d,i}^{\mathrm{p}}$, $C_{d,i}^{\mathrm{n}}$, $C_{s,i}^{\mathrm{p}}$, and $C_{s,i}^{\mathrm{n}}$). 
Here, a negative image refers to an image with a person detection confidence score below 0.25,\footnote{The YOLOv8 pose model also reports a confidence score for person detection using the default detection threshold set to 0.25.} indicating either the absence of a person or severe occlusion, which can negatively impact 3D pose estimation.  
As shown, the confidence score distributions on the server using the xlarge model are more widely spread toward both ends compared with those on the device using the nano model.  
This is because the xlarge model of the server attributes higher confidence scores to positive images and lower confidence scores to negative images than the nano model of the device. 
This disparity in the confidence score distributions clearly indicates that the inference accuracy and MPJPE can be improved by leveraging the xlarge model of the server through cooperative inference, rather than performing inference exclusively on the device.

\begin{figure}[t]
  \begin{center}
    \hspace{-3mm}
    \subfigure[]{
      \includegraphics[width=4.3cm]{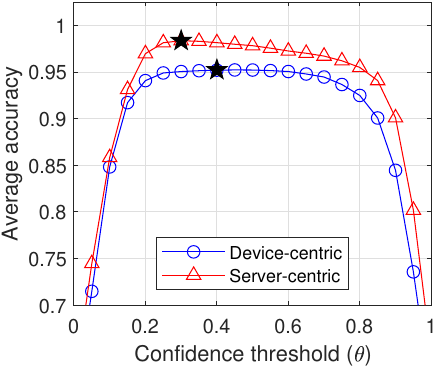}\label{graph2a}
    }
    \hspace{-3mm}
    \subfigure[]{
      \includegraphics[width=4.3cm]{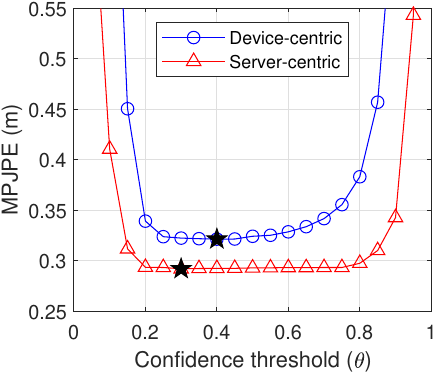}\label{graph2b}
    } 
  \end{center} 
  \vspace{-4mm}
  \caption{(a) Average accuracy and (b) MPJPE vs. confidence threshold ($\theta$) in the device- and server-centric inferences.}
\label{graph2}
\end{figure}
Fig. \ref{graph2} shows the average accuracy and MPJPE as functions of the confidence threshold in the device- and server-centric inferences. 
Consistent with the PDFs in Fig. \ref{graph1}, the server-centric inference using the xlarge model consistently achieves a higher accuracy and lower MPJPE than the device-centric inference using the nano model. 
Notably, an optimal confidence threshold exists that maximizes the accuracy and minimizes the MPJPE. 
This indicates that applying an optimal confidence threshold is essential for improving the accuracy and reducing the MPJPE.  
Furthermore, as demonstrated in Lemma \ref{lemma1}, the confidence threshold that maximizes the accuracy aligns with the threshold that minimizes the MPJPE for each inference method, suggesting that minimizing the MPJPE can be effectively achieved by maximizing the accuracy.

\begin{figure*}[t]
  \begin{center}
    \hspace{-3mm}
    \subfigure[]{
      \includegraphics[height=5cm]{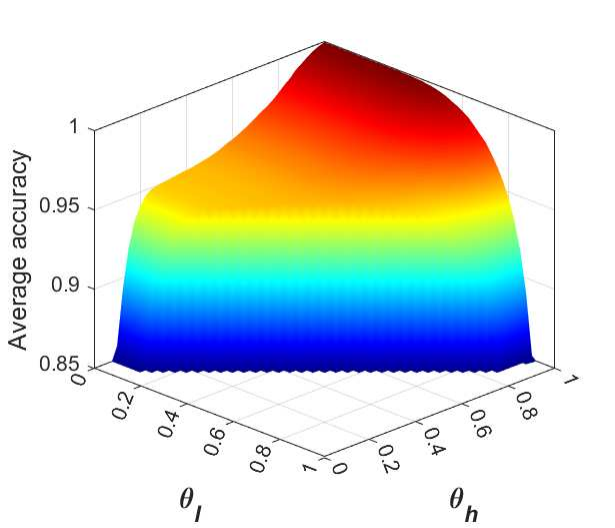}\label{graph3a}
    }
    \hspace{-3mm}
    \subfigure[]{
      \includegraphics[height=5cm]{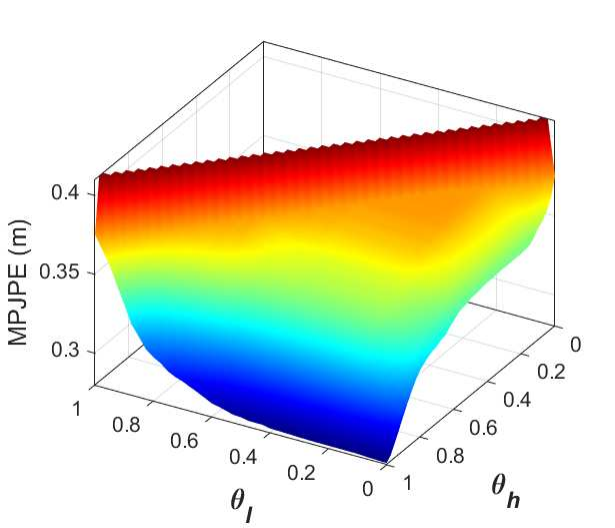}\label{graph3b}
    }
    \hspace{-3mm}
    \subfigure[]{
      \includegraphics[height=5cm]{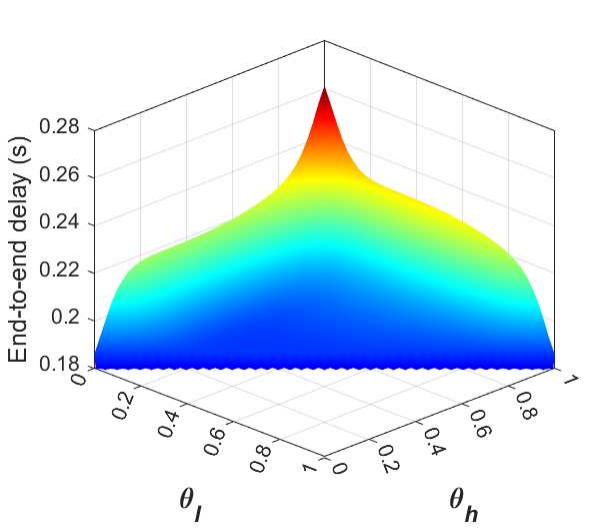}\label{graph3c}
    }
  \end{center} 
  \vspace{-4mm}
  \caption{(a) Average accuracy, (b) MPJPE, and (c) end-to-end delay vs. $\pmb{\theta}_{l}$ and $\pmb{\theta}_{h}$ when $\pmb{\theta}_{s} = 0.3$ in the proposed cooperative inference method.}
\label{graph3}
\end{figure*}

\begin{figure*}[t]
  \begin{center}
    \subfigure[]{
      \includegraphics[height=4.6cm]{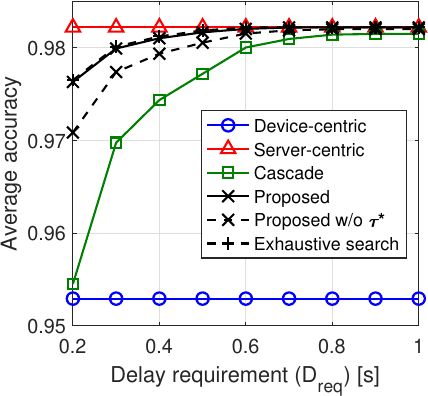}\label{graph4a}
    }
    \hspace{5mm}
    \subfigure[]{
      \includegraphics[height=4.6cm]{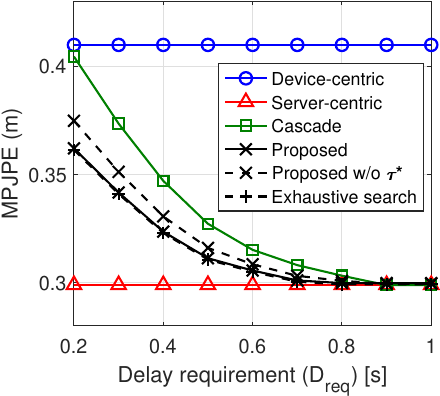}\label{graph4b}
    }
    \hspace{5mm}
    \subfigure[]{
      \includegraphics[height=4.6cm]{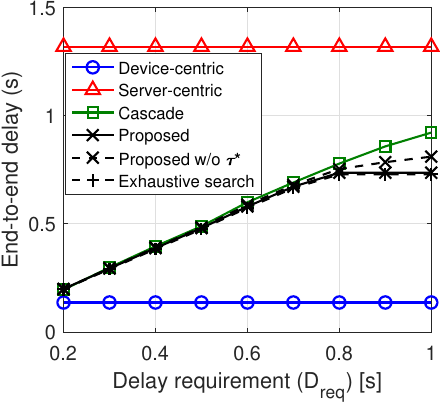}\label{graph4c}
    }
  \end{center} 
  \vspace{-4mm}
  \caption{Comparison of inference methods: (a) average accuracy, (b) MPJPE, and (c) end-to-end delay vs. delay requirement ($D_{req}$).}
\label{graph4}
\end{figure*}

Fig. \ref{graph3} shows the trends in average accuracy, MPJPE, and end-to-end delay according to the variations in the confidence thresholds at the end devices ($\pmb{\theta}_{l}$ and $\pmb{\theta}_{h}$) when the confidence threshold at the edge server ($\pmb{\theta}_{s}$) is fixed at 0.3 in the proposed cooperative inference method. 
When $\pmb{\theta}_{l}$ decreases and $\pmb{\theta}_{h}$ increases, the accuracy improves and MPJPE decreases.
This is because a wider interval between $\pmb{\theta}_{l}$ and $\pmb{\theta}_{h}$ results in more images being offloaded to the server, thereby enabling more accurate inference and 3D pose estimation through the larger NN model of the server. 
Instead, the end-to-end delay increases as $\pmb{\theta}_{l}$ decreases and $\pmb{\theta}_{h}$ increases due to the increase in offloaded data traffic.
Therefore, it is evident that changes in the two thresholds of the device create a trade-off between accuracy/MPJPE and end-to-end delay in the proposed cooperative inference.

Fig. \ref{graph4} compares the inference methods in terms of the average accuracy, MPJPE, and end-to-end delay as a function of the delay requirement ($D_{req}$).
The performances of device- and server-centric inferences are deterministic, regardless of $D_{req}$, and remain constant despite changes in its value.  
Device-centric inference achieves the shortest end-to-end delay and consistently satisfies the delay requirements but exhibits the lowest accuracy and highest MPJPE. 
In contrast, server-centric inference achieves the highest accuracy and lowest MPJPE but incurs the longest end-to-end delay and fails to satisfy the delay requirements.
Namely, the device- and server-centric inference methods exhibit a trade-off in performance. 
On the other hand, the cascade and proposed cooperative inferences demonstrate a balanced performance between device- and server-centric inferences. 
As $D_{req}$ increases, their accuracy improves and MPJPE decreases, thereby approaching the performance of server-centric inference.
This is because a more relaxed delay requirement allows more images to be offloaded to the server for inference.  
Accordingly, the end-to-end delays of both cascade and cooperative inferences gradually increase but always remain within $D_{req}$.
Eventually, they become saturated when $D_{req}$ exceeds a certain value.
This saturation occurs because sufficient offloading has already been completed to minimize the MPJPE.
Consequently, the proposed cooperative inference method exhibits significantly lower end-to-end delay while achieving the same accuracy and MPJPE as the server-centric inference.

Furthermore, the proposed inference consistently achieves better performance than cascade inference.
This is because, unlike cascade inference, which uses a single confidence threshold on the device, the proposed inference employs two thresholds, which enables more effective filtering of uncertain images that are re-inferred on the server, ultimately enhancing the accuracy and reducing MPJPE.
Moreover, when examining the performance of the proposed inference without the optimal transmission time ($\pmb{\tau}^{*}$) allocation, a slight degradation is observed compared with the optimized case. 
This result indicates that in a multi-device environment, optimizing resource allocation among devices contributes to an overall performance improvement.
Additionally, when comparing the performance of the proposed inference with that obtained through an exhaustive search for optimal parameters, only a negligible difference is observed, whereas the computational complexity remains significantly lower. In this experiment, the proposed algorithm requires an average of 0.015 s to obtain a single point, whereas the exhaustive search requires over 300 s on average.
This demonstrates that the proposed optimization algorithm effectively determines the optimal parameters within a reasonable time.

\begin{figure}[t]
  \begin{center}
    \hspace{-3mm}
    \subfigure[]{
      \includegraphics[width=4.3cm]{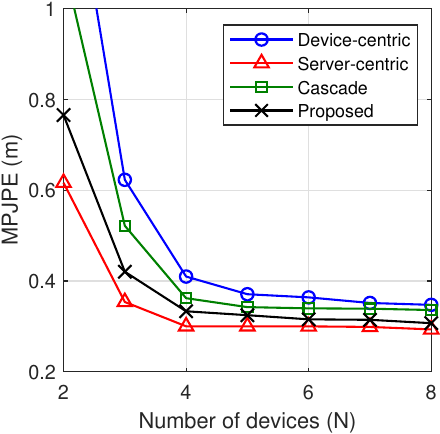}\label{graph5a}
    }
    \hspace{-3mm}
    \subfigure[]{
      \includegraphics[width=4.3cm]{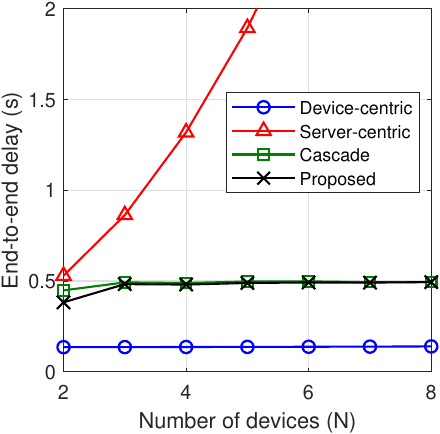}\label{graph5b}
    } 
  \end{center} 
  \vspace{-4mm}
  \caption{(a) MPJPE and (b) end-to-end delay vs. number of devices ($N$).}
\label{graph5}
\end{figure}

Fig. \ref{graph5} depicts the MPJPE and end-to-end delay versus the number of devices ($N$).
As $N$ increases, the MPJPE decreases sharply for all methods initially, but eventually reaches saturation.
It is obvious that increasing the number of devices effectively reduces the MPJPE up to a certain number; however, beyond that, additional devices contribute little to further reductions in the MPJPE.
The results suggest that $N=4$ is an appropriate number of devices for 3D pose estimation. 
Moreover, the increase in end-to-end delay for device-centric inference is negligible as $N$ increases, whereas server-centric inference shows a significant increase in end-to-end delay. 
This is because in the case of server-centric inference, the traffic volume of images transmitted to the server increases proportionally with the number of devices.
Even server-centric inference with a minimum value of $N=2$ fails to satisfy the delay requirement of 0.5 s. 
In contrast, both the cascade and proposed inferences always maintain end-to-end delays within this delay requirement.
This result indicates that the proposed inference method achieves an MPJPE comparable to that of the server-centric inference while satisfying the delay requirement.

\begin{figure}[t]
  \begin{center}
    \hspace{-3mm}
    \subfigure[]{
      \includegraphics[width=4.3cm]{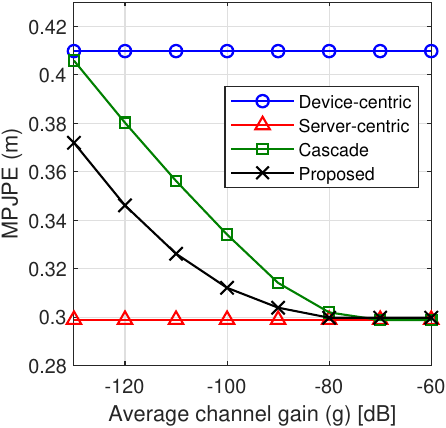}\label{graph6a}
    }
    \hspace{-3mm}
    \subfigure[]{
      \includegraphics[width=4.3cm]{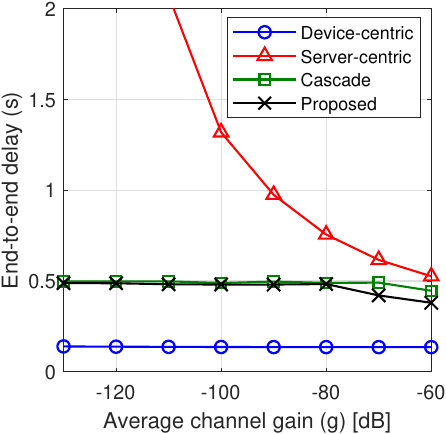}\label{graph6b}
    } 
  \end{center} 
  \vspace{-4mm}
  \caption{(a) MPJPE and (b) end-to-end delay vs. average channel gain ($g$).}
\label{graph6}
\end{figure}

Fig. \ref{graph6} shows the MPJPE and end-to-end delay versus average channel gain ($g$). 
The device-centric inference exhibits consistent MPJPE and delay performance regardless of $g$, whereas the server-centric inference shows a significant reduction in end-to-end delay as $g$ increases because of the decrease in the wireless transmission time ($T_{db}^{\mathrm{tx}}$).
For both the cascade and proposed inferences, an increase in $g$ allows more images to be offloaded to the server within the delay bound, resulting in a reduction in the MPJPE while ensuring $D_{req}$ = 0.5 s. 
Even when $g \geq -80$ dB, the proposed inference achieves the same MPJPE as the server-centric inference while providing lower end-to-end delay than $D_{req}$.
However, the server-centric inference fails to satisfy the delay requirement even when $g$ increases up to $-60$ dB. 

\begin{figure}[t]
  \begin{center}
    \hspace{-3mm}
    \subfigure[]{
      \includegraphics[width=4.3cm]{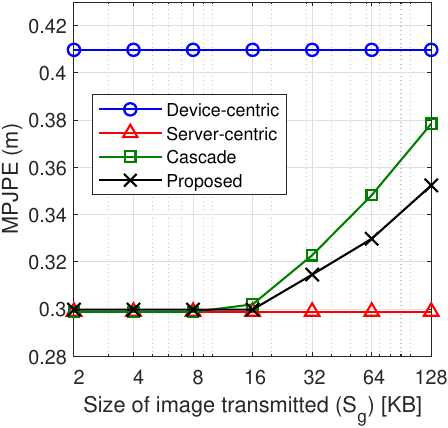}\label{graph7a}
    }
    \hspace{-3mm}
    \subfigure[]{
      \includegraphics[width=4.3cm]{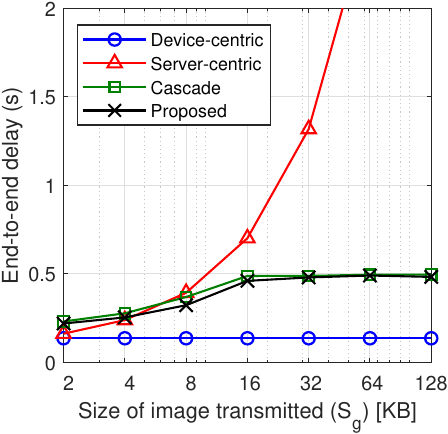}\label{graph7b}
    } 
  \end{center} 
  \vspace{-4mm}
  \caption{(a) MPJPE and (b) end-to-end delay vs. size of image transmitted ($S_{g}$).}
\label{graph7}
\end{figure}

Fig. \ref{graph7} illustrates the MPJPE and end-to-end delay versus the size of the transmitted image ($S_{g}$). 
As $S_{g}$ increases, the MPJPE of both the cascade and proposed inferences increases, and the end-to-end delay increases but remains bounded within $D_{req}$ = 0.5 s. 
In contrast, the server-centric inference experiences a significant increase in the end-to-end delay as $S_{g}$ increases due to the increase in the offloading data size. 
Only when $S_{g}$ is extremely small (below 4 KB), the end-to-end delay of the server-centric inference becomes shorter than that of the proposed inference, as the wireless transmission time ($T_{db}^{\mathrm{tx}}$) significantly decreases. 
This implies that server-centric inference can provide benefits over other methods only when the transmitted image size is substantially reduced by data compression or encoding techniques.

\begin{figure}[t]
  \begin{center}
    \hspace{-3mm}
    \subfigure[]{
      \includegraphics[width=4.3cm]{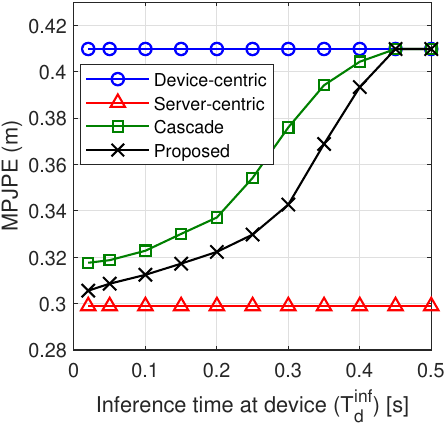}\label{graph8a}
    }
    \hspace{-3mm}
    \subfigure[]{
      \includegraphics[width=4.3cm]{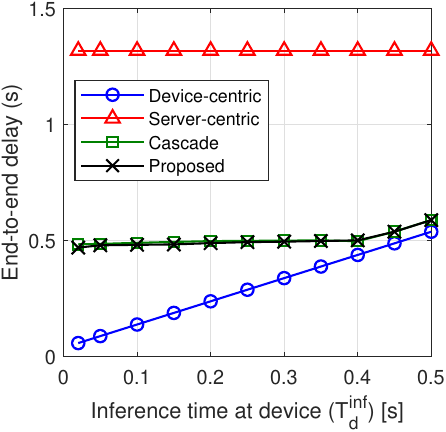}\label{graph8b}
    } 
  \end{center} 
  \vspace{-4mm}
  \caption{(a) MPJPE and (b) end-to-end delay vs. inference time at device ($T_{d}^{\mathrm{inf}}$).}
\label{graph8}
\end{figure}

Fig. \ref{graph8} shows the MPJPE and end-to-end delay versus the inference time at the device ($T_{d}^{\mathrm{inf}}$), which varies depending on the processor used in the device. 
For the cascade and proposed inferences, an increase in $T_{d}^{\mathrm{inf}}$ reduces the time margin available for offloading to the server, which in turn causes the MPJPE to increase. 
From the perspective of end-to-end delay, device-centric inference exhibits a proportional increase in delay as $T_{d}^{\mathrm{inf}}$ increases. 
Moreover, the proposed inference satisfies the delay requirement when $T_{d}^{\mathrm{inf}} \leq 0.4$ s. 
However, when $T_{d}^{\mathrm{inf}}$ exceeds this value, it fails to satisfy the delay requirement because $T_{d}^{\mathrm{inf}}$ consumes an excessively large portion of the 0.5 s delay bound.
Consequently, the proposed inference can no longer offload images to the server, and its MPJPE performance subsequently approaches that of device-centric inference. 
This implies that the proposed cooperative inference method is effective when $T_{d}^{\mathrm{inf}}$ is sufficiently smaller than $D_{req}$.


\section{Conclusion}
\label{title:con}
This study proposed a novel cooperative inference framework for real-time 3D human pose estimation in MEC networks, in which end devices with a smaller NN model collaborate with an edge server equipped with a larger NN model. 
A joint optimization problem was formulated to minimize the MPJPE while satisfying the end-to-end delay constraint, and optimal confidence thresholds and transmission times were derived using the proposed low-complexity optimization algorithm.
The experimental results demonstrated a trade-off between the MPJPE and end-to-end delay, depending on the confidence thresholds.   
Furthermore, they verified that the proposed cooperative inference method, with the optimal selection of confidence thresholds and transmission times, significantly reduced the MPJPE while guaranteeing the end-to-end delay requirement under various environmental factors, such as the number of devices, wireless channel gain, transmitted data size, and inference time at the device.
It is expected that the proposed cooperative inference method and optimization framework can be effectively applied to realize real-time 3D pose estimation in future MEC networks.
In future research, this study will be extended to real-world environments and multi-person scenarios to validate the practical applicability of the proposed approach.



\end{document}